\documentclass[letterpaper, 10 pt, conference]{ieeeconf}  % Comment this line out
\IEEEoverridecommandlockouts                              % This command is only
\usepackage[colorlinks,linkcolor=blue]{hyperref}
\usepackage{hyperref}
\usepackage{graphicx}
\usepackage{float}
\usepackage{multicol, blindtext}
\usepackage{caption}
\usepackage{subcaption}
\usepackage{adjustbox}
\usepackage{listings}
\usepackage{color}

\usepackage{comment}
 
\definecolor{codegreen}{rgb}{0,0.6,0}
\definecolor{codegray}{rgb}{0.5,0.5,0.5}
\definecolor{codepurple}{rgb}{0.58,0,0.82}
\definecolor{backcolour}{rgb}{0.95,0.95,0.92}
 
\lstdefinestyle{mystyle}{
    backgroundcolor=\color{backcolour},   
    commentstyle=\color{codegreen},
    keywordstyle=\color{magenta},
    numberstyle=\tiny\color{codegray},
    stringstyle=\color{codepurple},
    basicstyle=\footnotesize,
    breakatwhitespace=false,         
    breaklines=true,                 
    captionpos=b,                    
    keepspaces=true,                 
    numbers=left,                    
    numbersep=5pt,                  
    showspaces=false,                
    showstringspaces=false,
    showtabs=false,                  
    tabsize=2
}
 
\title{\LARGE \bf
Artificial neural networks condensation: A strategy to facilitate adaption of machine learning in medical settings by reducing computational burden
}

\author{Dianbo Liu$^{1,4,5}$ Nestor Sepulveda$^{2}$ and Ming Zheng$^{3}$% <-this % stops a space
\thanks{*}% <-this % stops a space
\thanks{$^{1}$Computer Science and Artificial Intelligence Laboratory, MIT, Cambridge, Massachusetts, 02139, United States
        {\tt\small dianbo at mit.edu}}%
\thanks{$^{2}$Department of Nuclear Science and Engineering, MIT, Cambridge, Massachusetts, 02139, United States
        {\tt\small nsep at mit.edu}}%
\thanks{$^{3}$Department of Physics, MIT, Cambridge, Massachusetts, 02139, United States
        {\tt\small smile321 at mit.edu}}%
\thanks{$^{4}$Harvard Medical School, Boston, Massachusetts, 02215, United States
        }%
\thanks{$^{4}$Boston Children's Hospital, Boston, Massachusetts, 02215, United States
        }%
}

\begin{document}

\maketitle

\thispagestyle{empty}
\pagestyle{empty}

%%%%%%%%%%%%%%%%%%%%%%%%%%%%%%%%%%%%%%%%%%%%%%%%%%%%%%%%%%%%%%%%%%%%%%%%%%%%%%%%
\begin{abstract}
Machine Learning (ML) applications on healthcare can have a great impact on people's lives helping deliver better and timely treatment to those in need. At the same time, medical data is usually big and sparse requiring important computational resources. Although it might not be a problem for wide-adoption of ML tools in developed nations, availability of computational resource can very well be limited in third-world nations. This can prevent the less favored people from benefiting of the advancement in ML applications for healthcare. In this project we explored methods to increase computational efficiency of ML algorithms, in particular Artificial Neural Nets (NN), while not compromising the accuracy of the predicted results. We used in-hospital mortality prediction as our case analysis based on the MIMIC III publicly available dataset. We explored three methods on two different NN architectures. We reduced the size of recurrent neural net (RNN) and dense neural net (DNN) by applying pruning of ``unused'' neurons. Additionally, we modified the RNN structure by adding a hidden-layer to the LSTM cell allowing to use less recurrent layers for the model. Finally, we implemented quantization on DNN forcing the weights to be 8-bits instead of 32-bits. We found that all our methods increased computational efficiency without compromising accuracy and some of them even achieved higher accuracy than the pre-condensed baseline models. %The hidden-layer LSTM model which doubled the training speed and the quantized DNN which cut the memory usage by 90\% yield the benchmark condensation performance.

\end{abstract}

%%%%%%%%%%%%%%%%%%%%%%%%%%%%%%%%%%%%%%%%%%%%%%%%%%%%%%%%%%%%%%%%%%%%%%%%%%%%%%%%
\section{Introduction}
Machine Learning (ML) applications on health care can have a great impact on people's lives. Today, the possibilities for ML in health care include diagnostic systems, biochemical analysis, image analysis and drug development among others. One of the most significant challenges in using ML for health care applications is that data is usually huge and sparse, requiring important computational resources. In consequence, the availability of computational resources to utilize such tools can limit their wide-spread use. For developed countries finding a computer that can run ML algorithms might not be a problem. People living in developed countries seldom consider the efficiency of the computational model they build, since it is always easy for them to access the devices with cutting-edge performance where overparameterization is not a big deal as along as overfitting is treated appropriately. However, people living in the third world might not have access to the computational resources of the same level. In consequence, increasing computational resources efficiency in ML implementations can have a significant impact their adaption and a significant impact on their lives.

One medical application of ML is the use of neural networks (NNs) to predict the mortality of a patient transferred into the intensive care unit (ICU), based on his/her vital signs, laboratory tests, demographics, and etc. Mortality prediction is important in clinical settings because such a prediction can help determine the declining state and need for intervention. The nature of the data use for in-hospital mortality prediction is sequential as it is generated from the beginning of the stay of a patient and over the time of the stay. Labels are simple, since the result for each patient is to live the ICU (0) or deceased (1). Additionally, sample data  for research purposes is readily available at the MIMIC-III database. For the previous reasons we use in-hospital mortality prediction as our application of ML for our study.

In this study, we will explore ways to improve efficiency (training speed, memory size, inference speed) of ML algorithms based on NNs for in-hospital mortality prediction. Our goal is to find ways in which without compromising prediction accuracy models can be more efficient.  We explored efficiency improvements in Recurrent Neural Nets (RNN) and Dense Neural Nets (DNN) architectures without losing the prediction accuracy, so that our models could potentially be trained and run on the slower devices. Moreover, we are the first pioneers whoever try to do RNN pruning with clinical implementation.

Reduction of complexity and improvement of efficiency of artificial neural networks is an active research field. A wide range of methods have been explored. One representative example is neural network pruning,where a fraction of weights are removed from the trained model  and the ``lottery ticket'' is found when the remained weight can still be quickly trained with competitive loss and accuracy\,\cite{Frankle2018,See2016,Han2015,Babaeizadeh2016}. Another method is embedding more information into the microcosmic units by inserting deep neural networks (DNN). In \cite{Dai2018}, DNNs were inserted between the recurrent layer and the input (masking) layer for each gate in the LSTM to form a LSTM embedded with hidden layers (hLSTM). By modifying the microcosmic architecture (i.e. the original LSTM cell), they were able to simplify the macroscopic architecture--say, reducing the number of LSTM layers--to achieve the more efficient setup (e.g. fewer number of total parameters, faster training speed, etc.). Another post-training condensation method called quantization, where parameters originally stored in 32 bits floating point format were forcely converted to 8 fixed bits\cite{Krishnamoorthi2018}. Other methods used for neural network condensation include but are not limited to binarization of neural networks\cite{Courbariaux2016}, knowledge distillation\cite{Hinton2016} and Huffman coding \cite{Han2015}. In this study, we used hLSTM, neural network pruning and quantization to condense sizes of neural networks, increase speed while maintaining their predictive accuracy. 

\section{Materials and Methods}

\subsection{Intensive care unit data}
We used \href{https://mimic.physionet.org/about/mimic/}{MIMIC-III} critical care database for the implementation of our models\,\cite{Johnson2016}.
53,423 distinct hospital adult patients admitted to critical care units between 2001 and 2012 are included in this database. We excluded all neonatal and pediatric patients (age 18 or younger at time of ICU stay) because  the physiology of pediatric critical care patients differs significantly from adults \cite{Harutyunyan2017}. We also excluded any hospital admissions with multiple ICU stays or transfers between different ICU units. The final cohort has 33,798 unique patients with a total of 42,276 hospital admissions and ICU stay. We define a test set of 5,070 (15\%) patient stays. In-hospital mortality is determined by comparing patient date of death (DOD) with hospital admission and discharge times. The mortality rate within the cohort is 11\%. The median age of adult patients is 65.8 and 55.9\% patients are male. A mean of 4579 charted observations and 380 laboratory measurements as well as other static information are available for each hospital admission. 
%This is a public open data set with good maintaining as well as user-friendly downloading, upgrading and corrections. Basically, the data quantity is sufficient and reliability is high enough.

For our purpose, 76 features are selected for our research. Those features are listed in Table.\ref{tab:feature}. Some features may appear for multiple times (in different means or conditions) thus are regarded as independent features. More details about data processing are referred to \emph{Data prepossessing}.

\begin{table*}[h]
   \centering
      \caption{Feature selection.}
   %\topcaption{Table captions are better up top} % requires the topcapt package
    \begin{tabular}{cccc} % Column formatting, @{} suppresses leading/trailing space
      \hline\hline
      pH &  Fraction inspired oxygen &  Systolic blood pressure & Height\\
      Weight & Oxygen saturation & Diastolic blood pressure & Glucose  \\
      Temperature & Mean blood pressure & Capillary refill rate  &  Respiratory rate\\
      Heart Rate & Fraction inspired oxygen & [Glascow coma scale]$\times$50  & \\
    \hline
  \end{tabular}
   \label{tab:feature}
\end{table*}

\subsection{Data prepossessing}

The data was taken from MIMIC-III database. Only the first 48 hours are used in our inputs (take around an hour) and 76 features are selected for our research. Those features are listed in Table.\ref{tab:feature}. We resampled the time series into regularly spaced intervals. If there were multiple measurements of the same variable in the same interval, we used the value of the last measurement. We imputed the missing values using the previous value if it exists and a pre-specified ``normal" \cite{Harutyunyan2017} value otherwise. In addition, we add a binary mask input for each variable indicating the time steps that contain a true (vs. imputed) measurement \cite{Zachary2016}. Categorical variables were encoded using a one-hot vector at each time step. Then the inputs are then normalized by subtracting the mean and dividing by standard deviation. Statistics were calculated per variable after imputation of missing values.

Since there is no direct access to the MIMIC-III dataset unless passing the Collaborative Institutional Training Initiative (CITI Program), we have our team members trained for a whole day to get the certification %(Fig.\,\ref{fig:CITI}). 

\begin{comment}
\begin{figure}[htbp]
\centering
   \includegraphics[width=0.7\linewidth]{CITI.pdf} % requires the graphicx package
   \caption{{\bf  Certification for capable of downloading MIMIC-III dataset.} }
   \label{fig:CITI}
\end{figure}
\end{comment}

\subsection{Computational environment}
Python3.6, keras 2.2.4 using tensorflow 1.1.2 as backend was used for analysis in this study. 

\subsection{Performance metrics}
Classification accuracy of all models were measured in AUROC (also called AUCROC) on test set. Sizes of model were measured by number of parameters and sizes of the saved model file. Inference speed was calculated based on time taken to making prediction on test data and were normalized to per patient. 

%%%%%%%%%%%%%%%%%%%%%%%%%%%%%%%%%%%%%%%%%%%%%%%%%%%%%%%%%%%%%%%%%%%%%%%%%%%%%%%%

\begin{figure*}[htbp]
\centering
   \includegraphics[width=0.7\linewidth]{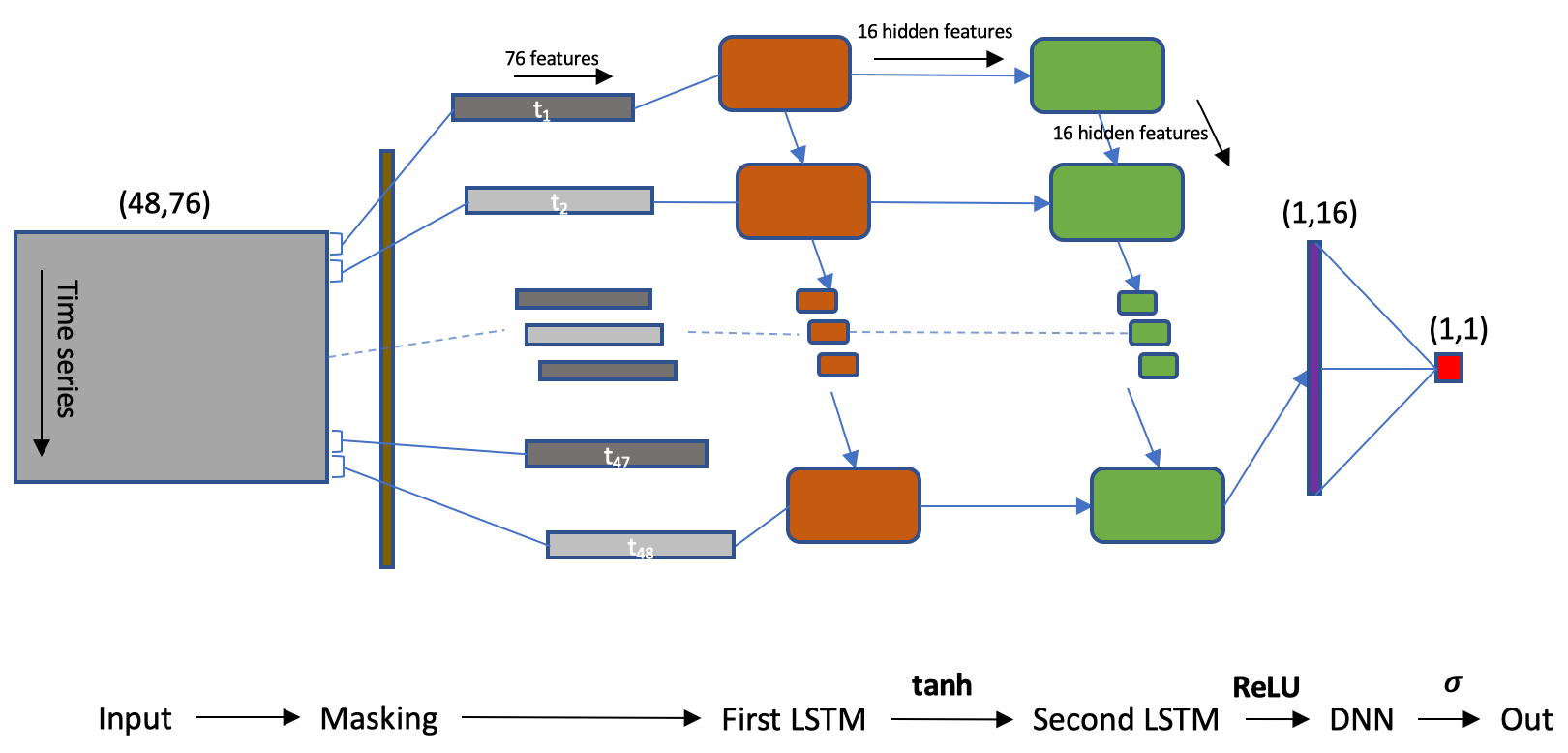} % requires the graphicx package
   \caption{{\bf  Architecture of RNN baseline model.} }
   \label{fig:baseline}
\end{figure*}

\subsection{Recurrent neural network model}
Our RNN baseline model is designed as a recurrent neural network consisting of a masking layer, two LSTM layers, a dropout layer and a dense output layer as shown in Fig.\,\ref{fig:baseline}. We chose two layers of LSTM because during our literature review we identified this structure as the one with best performance in the MIMIC-III mortality prediction work we found\,\cite{Harutyunyan2017}. The masking layer masks (skips) the time step for all downstream layers if the values of input tensor at the time step all equal to zero, which represents missing data for that time step. The first layer of LSTM takes in the original 76 features and generates a 16-feature hidden state based on the hidden state of the previous step and the new coming observation, then such a hidden state is forward to the entrance of the second LSTM layer, which produces another 16-feature hidden state at each step. A dropout layer is followed by the last-step hidden state of the second LSTM layer to prevent complex co-adaptations of the neurons. Finally, a dense layer is used to generate a soft 0/1 mortality prediction.

\subsection{Hidden-layer LSTM}
Besides pruning upon RNN, we also tried another way---inserting an additional hidden dense layer into the inner gates LSTM (let's call it hLSTM) to improve the ``power'' of the LSTM. For a traditional LSTM, we have the inner structure as following:
\begin{eqnarray}
\left( \begin{array}{c} f_t \\ i_t \\o_t \\ g_t \end{array} \right) &=& \left( \begin{array}{c} \sigma(W_f\ast[x_t,h_{t-1}]+b_f) \\ \sigma(W_i\ast[x_t,h_{t-1}]+b_i) \\ \sigma(W_o\ast[x_t,h_{t-1}]+b_o) \\ \mathrm{tanh}(W_g\ast[x_t,h_{t-1}]+b_g) \end{array} \right), \label{eq:lstm_r} \\
c_t &=& f_t\otimes c_{t-1} + i_t \otimes g_t, \\
h_t &=& o_t \otimes \mathrm{tanh}(c_t),
\end{eqnarray}
where $\ast$ is the matrix product while $\otimes$ is the element-wise product. $W_:$ are recurrent kernel matrices of the gates and $b_:$ are corresponding bias terms. $f$, $i$, $o$, $c$, $x$, $h$ and $c$  the forget gate, input gate, output gate, vector for cell updates, input, hidden state, and cell state, respectively. Subscript $t$ indicates the time step.
For hLSTM, the recurrent layer in Eq.\ref{eq:lstm_r} is modified as:
\begin{small}\begin{eqnarray}
\left( \begin{array}{c} f_t \\ i_t \\o_t \\ g_t \end{array} \right) = \left( \begin{array}{c} \sigma(W_f\ast\mathrm{ReLU}(H_f\ast[x_t,h_{t-1}])+b_f) \\ \sigma(W_i\ast\mathrm{ReLU}(H_i\ast[x_t,h_{t-1}])+b_i) \\ \sigma(W_o\ast\mathrm{ReLU}(H_o\ast[x_t,h_{t-1}])+b_o) \\ \mathrm{tanh}(W_g\ast\mathrm{ReLU}(H_g\ast[x_t,h_{t-1}])+b_g) \end{array} \right).\nonumber
\end{eqnarray}\end{small}
As we can see, $[x_t, h_{t-1}]$ is processed with a DNN $H$ before matrix multiplication with $W$.

\subsection{Feedforward dense neural network}
Our baseline feed forward artificial neural network---commonly called the deep neural network (DNN)---used in this project consists of three fully connect layers , a dropout layers and an output layer. The fully connect layers have 256, 128 and 64 neurons respective and use ReLU as activation function. The dropout layer has a probabilistic dropout rate of 0.5. Sigmoid function was used as activation at the output layer. The loss function was binary cross-entropy  and the optimization algorithm was Adam. The baseline DNN model and the pruned DNN model (pDNN) were all trained for 20 epochs using a batch size of 8.

\subsection{Neural network pruning and quantization}
All the neural network prunings were conducted at channel level, which means a neuron and all its inputs and outputs were removed from the model if the neuron is pruned. Keras surgeon library in python (\url{https://github.com/jiajuns/Neural-Network-Pruning-Keras}) was used for the pruning. In each layer, neurons were pruned if they mean weight across all inputs from previous layer were below the set quantiles (50\% in this case). Quantization was applied on trained DNN after training. Parameters originally stored in 32 bits floating point format were converted to 8 bits using tensorflow lite.

%%%%%%%%%%%%%%%%%%%%%%%%%%%%%%%%%%%%%%%%%%%%%%%%%%%%%%%%%%%%%%%%%%%%%%%%%%%%%%%%

\begin{figure}[htbp]
\centering
    \begin{subfigure}{.95\linewidth}
     \centering
     \includegraphics[width=0.95\linewidth]{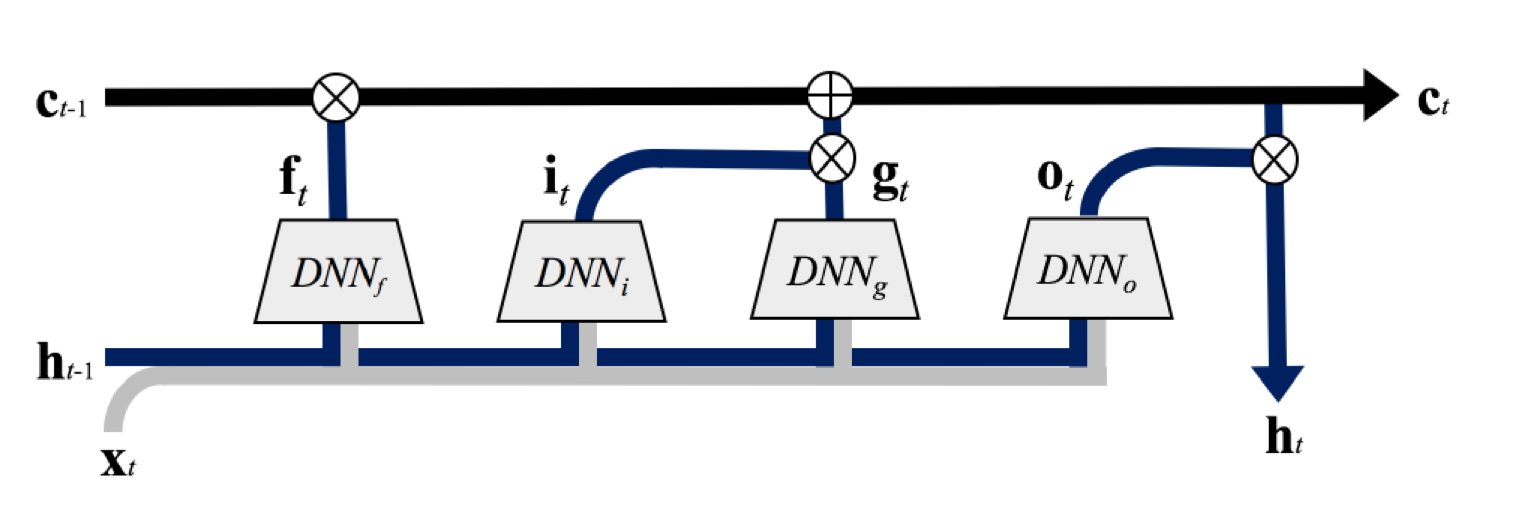}
     \caption{}
     \end{subfigure}
     \\
      \begin{subfigure}{.95\linewidth}
     \centering
     \includegraphics[width=0.8\linewidth]{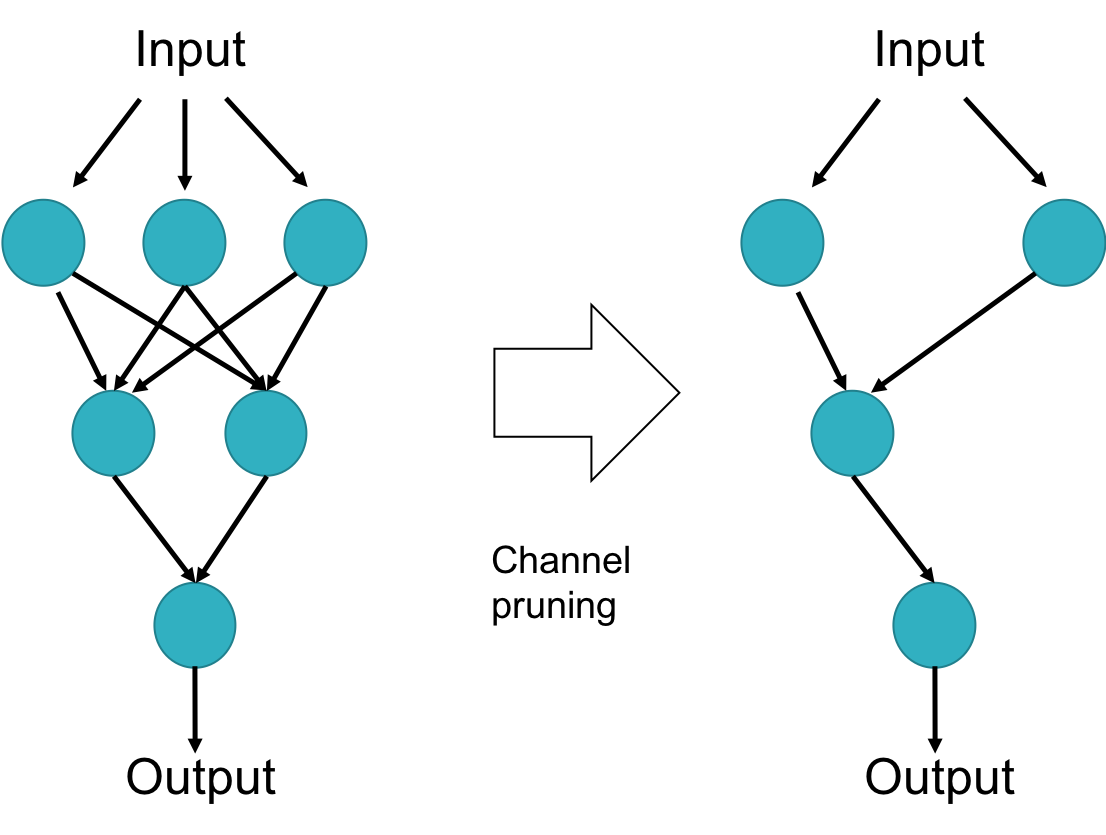}
     \caption{}
     \end{subfigure}
     \\
      \begin{subfigure}{.95\linewidth}
     \centering
     \includegraphics[width=0.95\linewidth]{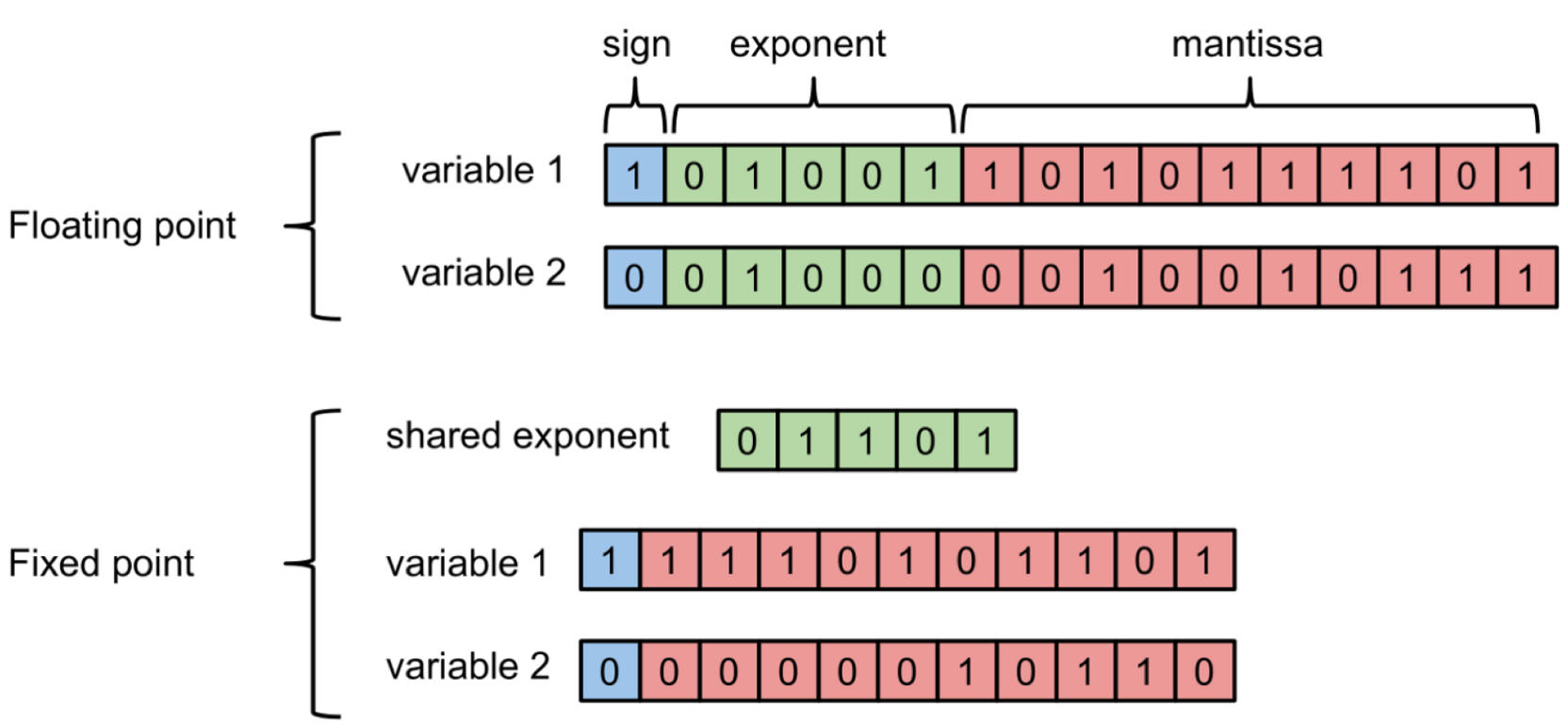}
     \caption{}
     \end{subfigure}
   \caption{{\bf Neural network condensation methods(a) Hidden layer LSTM.} Instead of single fix layer non-linearity for gate control of LSTM, multiple layer NN with ReLu as activation were used to enhance the gate controls. In this way, fewer layers of LSTMs were needed to build a model with similar performance. (b) A large portion of parameters in artificial neural networks are redundant. We pruned 50\% of the channels(neurons) with lowest weights in each layer to reduce size and complexity of the neural network. (c) Most artificial neuron network implementation in research settings uses 32 or 64 bit floating point for model parameters. We quantized the parameters to 8 bits after training to reduce sizes of the models.  }
   \label{fig:methods}
\end{figure}

\section{results}

%\subsection{Patient cohort}

\subsection{Recurrent network: hLSTM and pruned LSTM}
Recurrent artificial neural network neural network is a widely used machine learning model in clinical settings.We built a baseline RNN using two layers of Long Short-Term Memory neurons (LSTMs).After training, the RNN model achieved a decent performance of AUCROC=0.85 (Table \ref{tab:RNN}). In order to enable the machine learning algorithms to be used on devices with limited computational power such as those in developing countries, we used three strategies to reduce the storage size of the model and to increase speed of  training and inference (Figure \ref{fig:methods}).

% \begin{figure}[htbp]
% \centering
%   \includegraphics[scale=0.18]{hLSTM_code.png}
%   \caption{{\bf  Creating hLSTM in Keras documents.} The upper panel is the original Keras LSTM, where recurrent kernel is applied to the reduced observations and hidden states immediately. The lower panel is the modified LSTM (hLSTM), where $x$ and $h$ are processed with the hiden kernel before being forward to the recurrent activation.}
%   \label{fig:hLSTM_code}
% \end{figure}

The first strategy is to use modify LSTM neuron to increase representation power of the neuron. Since there is no direct hLSTM model available in Tensorflow, Keras, or Pytorch, We spent a lot of time trying our best to understand the designing principle of Keras and finally were able to successfully modified the original Keras code structure and added a additional hidden layer into the original LSTM class. The main modification are shown in Listing.\,\ref{fig:hLSTM_code}. Compared with the old LSTM, we can tell that one additional layer called ``hidden\_karnel'' is inserted between the input kernel and the recurrent kernel. By using this strategy, we replaced the old two-layer LSTM with only one layer of such a hLSTM, so that we simplified the overall structure by trying to embedding the same quantity of information in this single ``condensate'' layer.

\begin{table*}[ht]
   \centering
   \caption{Recurrent neural networks condensation}
   %\topcaption{Table captions are better up top} % requires the topcapt package
    \begin{tabular}{@{} c  | c | c | c | c | c @{}} % Column formatting, @{} suppresses leading/trailing space
      \hline%\hline
        ~ &Params & FileSize & Inference & Training time [s] & Test AUROC \\
      \textbf{Model} & [\# count] & $\mathrm{[Kb]}$ & [$\mu$s/sample]  & (20 epochs) & (last epoch) \\
      \hline
      Baseline LSTM & 8,081 & 129 & 523 & 4,890  & 0.836\\
      %\hline
      Pruned LSTM & 3,273 &  73 & 318  & 4,990 & 0.853\\
     % \hline
      Hidden-Layer LSTM & 6,993 & 111 & 254 & 3,000 & 0.860 \\
      %\hline
\hline
  \end{tabular}
   \label{tab:RNN}
\end{table*}

Both the baseline model and the new model with only one layer of hLSTM are trained under the same training settings (\emph{dropout}=0.3 and \emph{learn\_rate}=0.001). The comparison of AUROC and accuracy are shown in Fig.\,\ref{fig:performance}. The number of parameters of these two models are listed in Table.\ref{tab:RNN}. As we can see, this simplified model with only one layer of hLSTM beats the baseline model 2-fold in training speed, 32\% reduction in parameter numbers and maintains the higher AUROC at the same time.
\par 

An alternative to hLSTM for RNN model condensation is pruning. 50 \% of LSTM neurons with lowest weights in each hidden layers were pruned after first epoch of training. After training the pruned model was trained for another 19 epochs. The pruned LSTM only has half of the number parameters of original LSTM but achieved the similar level of accuracy, giving AUCROC of 0.85. The inference speed of pruned LSTM also doubled compared with the original LSTM (Table\ref{tab:RNN}).

%NOTE: ADD RESULTS ON PRUNED LSTM

\begin{figure}[htbp]
\centering
   \includegraphics[width=1\linewidth]{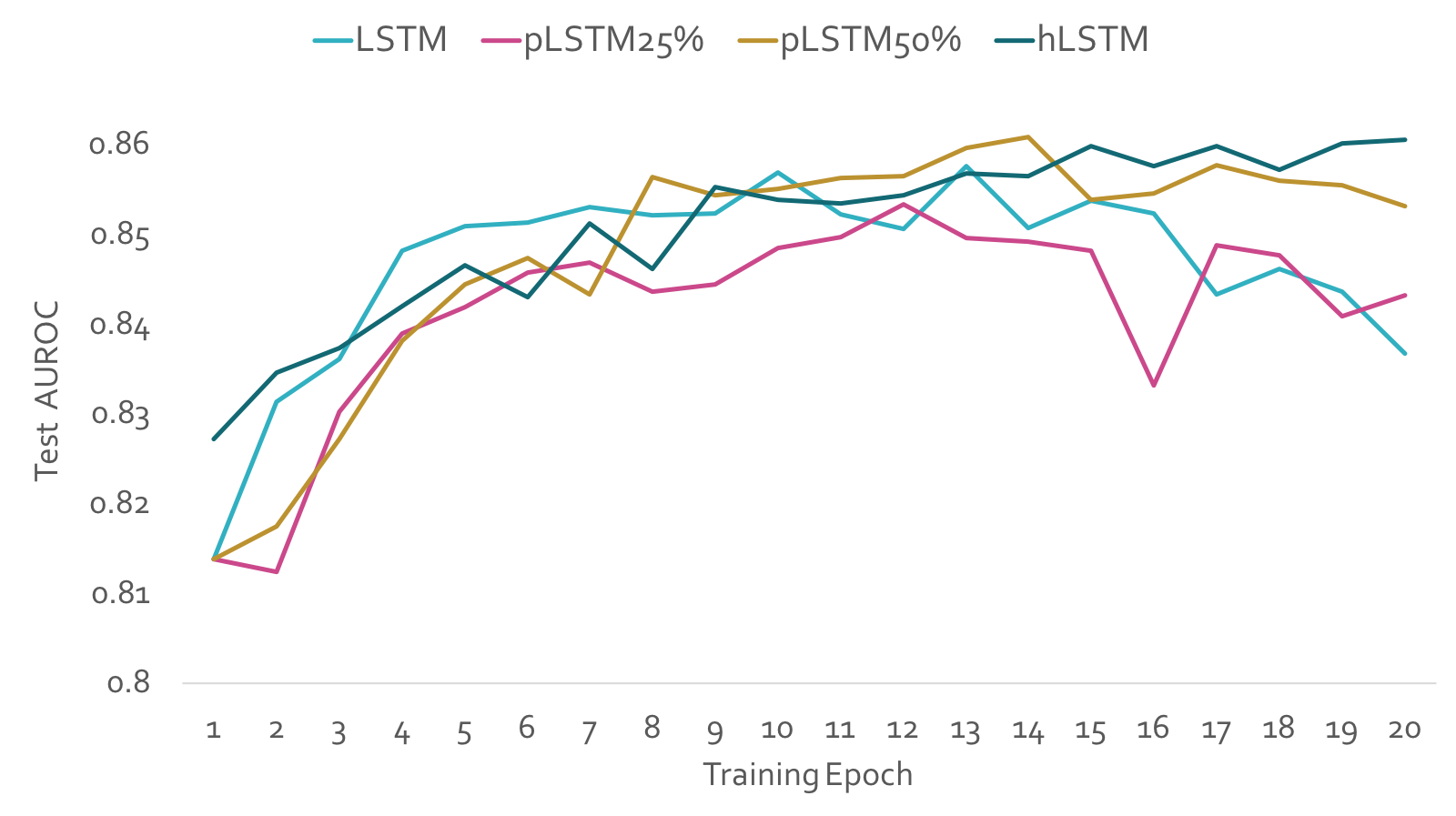}% requires the graphicx package
   \caption{{\bf Test AUROC by training epoch for RNN models.} Evolution of different RNN models over training epochs on test data. The percentage next to the pruned LSTM (pLSTM) model indicates the pruned percentile.}
   \label{fig:RNN_e}
\end{figure}

\subsection{Feedforward neural network: pruning and quantization}
\begin{figure}[htbp]
\centering
  \begin{subfigure}{.99\linewidth}
     \centering
     \includegraphics[width=0.99\linewidth]{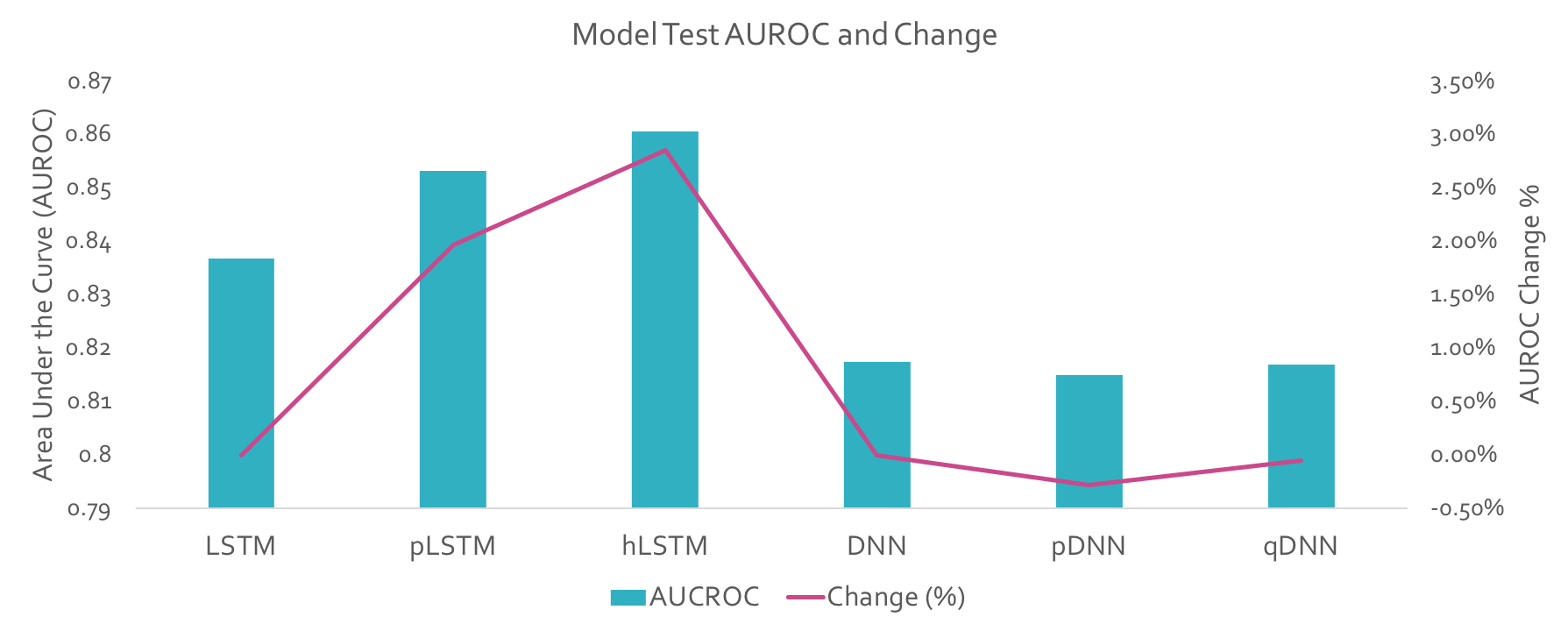}
     \caption{}
     \end{subfigure}
     \\
      \begin{subfigure}{.99\linewidth}
     \centering
     \includegraphics[width=0.99\linewidth]{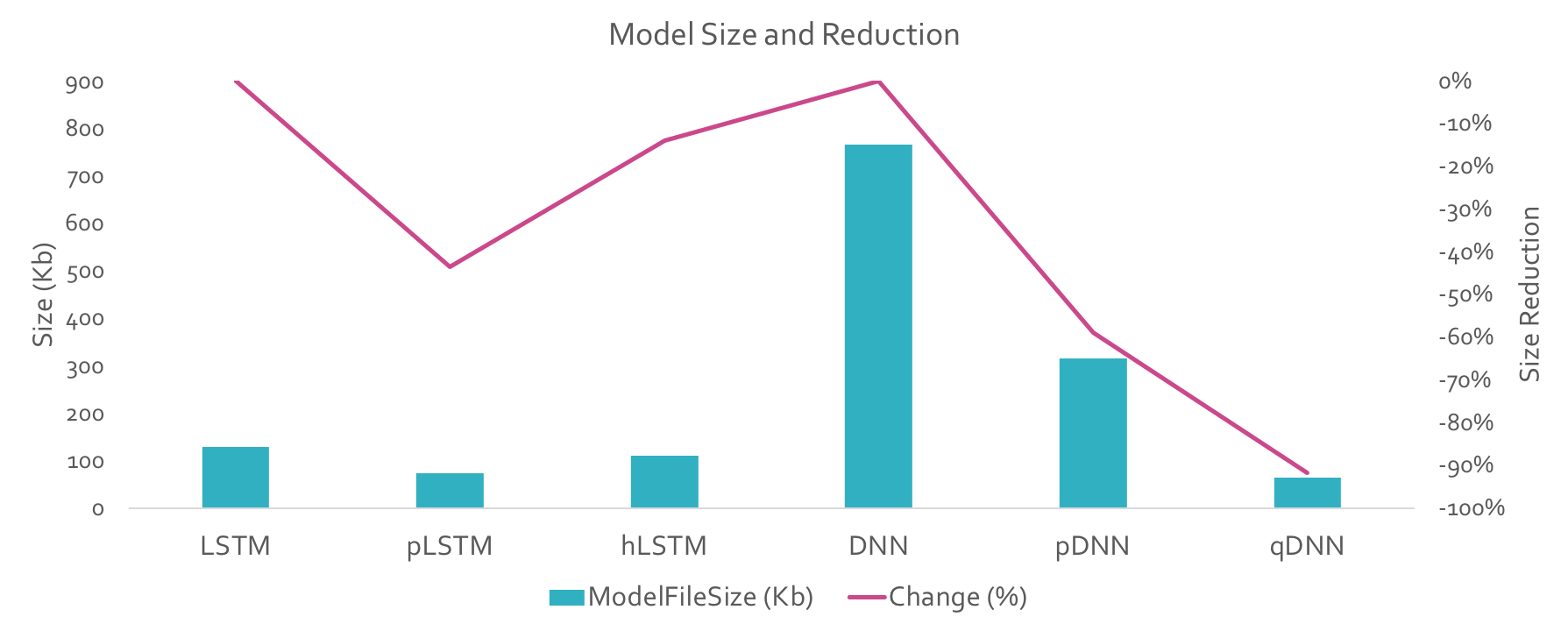}
     \caption{}
     \end{subfigure}
     \\
      \begin{subfigure}{.99\linewidth}
     \centering
     \includegraphics[width=0.99\linewidth]{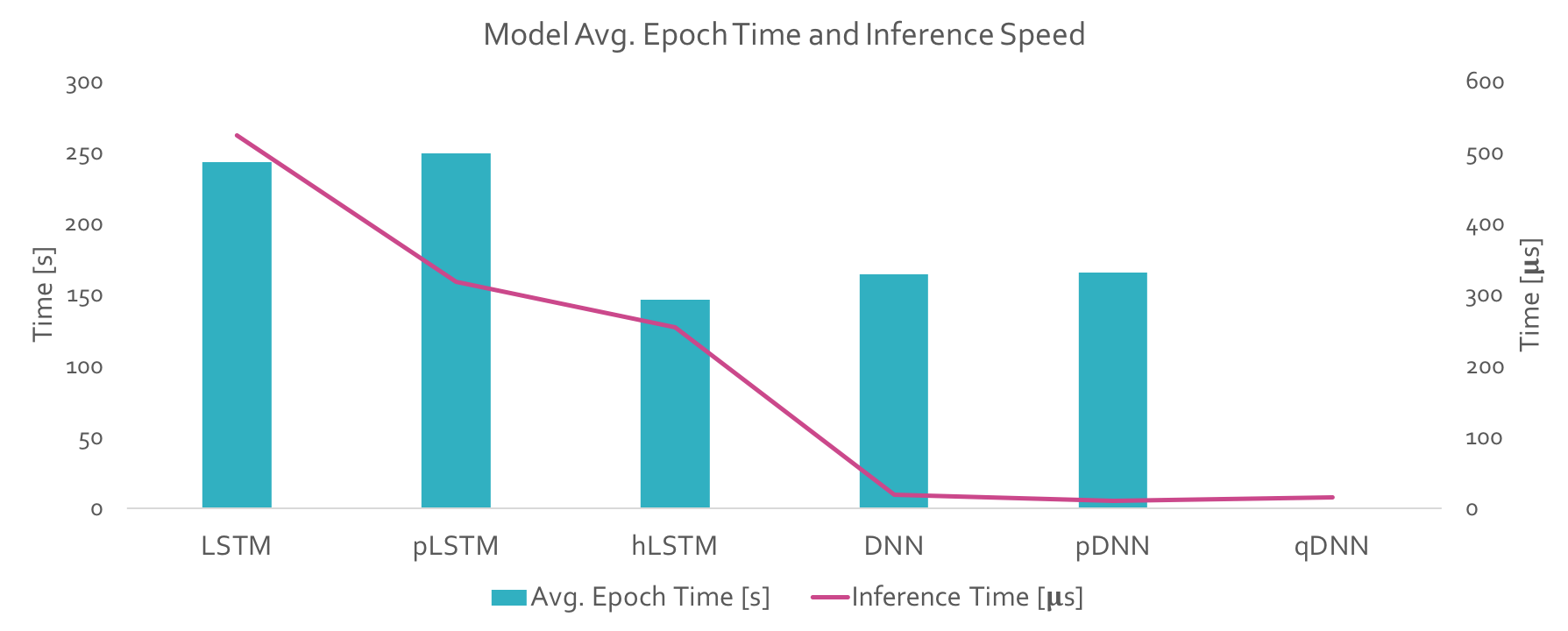}
     \caption{}
     \end{subfigure}
   \caption{{\bf Accuracy, model size and inference speed of recurrent and feedforward neural network after different types of condensation.} (a) AUCROC of various models. (b) various model sizes in memory. (c) inference speed of various models. LSTM: RNN baseline model with two layers of LSTM; pLSTM: pruned LSTM model; hLSTM: RNN model with one layer of hidden-layer inserted LSTM; DNN: DNN baseline model; pDNN: pruned DNN model; qDNN: quantized DNN model.}
   \label{fig:performance}
\end{figure}

Feedforward neural network or commonly called deep nerual network (DNN), if it has multiple hidden layers, is another widely used machine learning in clinical settings.The input into DNN model has the same feature set as LSTM model. The values were calculated by averaging non-missing values across time steps.  We trained DNN with 3 hidden layers , consisting of 256, 128 and 64 neuron at each layers, to make mortality prediction. The DNN achieved an AUROC of 0.82 using patients data of the first 48 hours after admission. We explored two method to condense the size of the DNN. The first method was pruning, using the pruning strategy as in RNN where 50\% of the channels were pruned after first epoch of training, the prediction accuracy of the pruned DNN (pDNN) maintained at the same level as the original DNN and the inference speed doubled (Table \ref{tab:DNN}). Quantization refers to the process of reducing the number of bits that represent a number. In the context of this project, the predominant numerical format used was 32-bit floating point. We used a after-training-quantization strategy to represent the parameters of the DNN model using 8-bit integers (qDNN). This method reduced storage size of the DNN model by 5 times without incurring significant loss in accuracy (Table \ref{tab:DNN}). We also compared the overall performances of DNN condensation with those of RNN in Fig.\,\ref{fig:performance}. 

\begin{table*}[ht]
   \centering
   %\topcaption{Table captions are better up top} % requires the topcapt package
   \caption{Feedforward neural networks condensation}
    \begin{tabular}{@{} c  | c | c | c | c | c @{}} % Column formatting, @{} suppresses leading/trailing space
      \hline%\hline
        ~ & Params & FileSize & Inference & Training time [s] & Test AUROC \\
      \textbf{Model} & [\# count] & $\mathrm{[Kb]}$ & [$\mu$s/sample]  & (20 epochs) & (last epoch) \\
      \hline
      Baseline DNN & 60,929 & 767 &20 &3,300 & 0.82\\
      %\hline
      Pruned DNN  & 27,312 &315 &10 & 3,310 &0.81\\
      %\hline
      Quantized DNN &60,929 &64& 15 &NA&0.82 \\
      %\hline
\hline
  \end{tabular}
   \label{tab:DNN}
\end{table*}

% following are the two KEY figures as Dianbo suggest.

\section{Discussion}
 In the present work we were able to use MIMIC III data to train in-hospital mortality neural net models with high accuracy. The models were then ``treated'' with different methods to gain efficiency (memory size reduction, increased speed) without compromising accuracy. Some of out treatments resulted in consistently higher accuracy models. We implemented state of art NN architectures for both Recurrent Neural Nets and Dense Neural Nets proving our methods add value in both settings. From the above, we conclude our efficiency treatments can be extended to other medical applications using similar data, and probably non-medical applications as well.

The implementations used in this work required us to write our own code not only for creating models, and processing data, but also to apply our methods. Moreover, we went as far as to modify the native LSTM implementation in Keras (python) in order perform our analysis. This ``exploration'' was only possible thanks to the concepts learned in 6.867 and helped us to gain a greater understanding on the implementation of ML models.

Our results show a great potential for our methods and particularly for hidden-layer LSTM models as they achieved higher accuracy with simplified architectures (with a doubled the training speed). The quantized DNN which cuts the memory usage by 90\% without losing test AUROC and has much higher inference speed then those RNN models is also a promising option when the capacity of the device is limited (e.g. the thrid-world devices or mobile devices). In future research we would like to explore the effects or using pruning or quantization on top of this type of model to further explore the possible gains.

\section{Literature review of artificial NN condensation methods}

To increase NNs' efficiency, the core task is to reducing the number of parameters without hurting the loss and accuracy. Generally speaking, There are two classes of approaches to do so: 1) removing parameters from a trained model, and 2) directly build NNs with compressed parameter representation or optimized the architecture. The approach 1) class is also known as network pruning. One simple way is to do one-shot pruning\,\cite{Frankle2018}, where each time a fraction of weights are removed from the trained model and the other weights are set back to the initialization value, and the ``lottery ticket'' is found when the remained weight can still be quickly trained with competitive loss and accuracy. Such a one-shot process can be iteratively done until some thresholds meet. Finally, they found Neural network compression techniques are able to reduce the number of parameters of trained neural networks by 90 percent. Same other simple ways including: ``Learning-Compression'' Algorithms\,\cite{Carreira2018}, pruning and splicing\,\cite{Guo2016}, Net-Trim\,\cite{Aghasi2017}, and $L_0$ regularization\,\cite{Louizos2017} where the parameters are penalized under some norm to generate a optimization problem; NoiseOut\,\cite{Babaeizadeh2016} where correlation between activation of neurons are computed by adding noise to outputs followed by back-propagation. Quantization\,\cite{Choi2018} is another pruning concept: they assume that redundant weights in deep neural network are distributed as several semi-continuous clusters, resulting in the possibility of quantizing those clusters by picking up the most representative weight and the do weight sharing among all neurons within the same cluster. There is another method called Fisher Pruning\,\cite{Theis2018}, where they assume the trained model is located at a local minimum of its loss; thus by estimating the Hessian matrix they would be able to know which direction gets the smallest estimated increase in loss and prune the weight that has the maximum gradient projection along this direction. There are also some fancy approaches where the author use another NN to learning and conduct the best pruning decisions upon the network to be pruned (the backbone NN): in \cite{Lin2017}, they developed a method called RNP to model their pruning process as a Markov decision process and use reinforcement learning for training via an additional RNN; in \cite{Zhong2018}, they used LSTM to guide an end-to-end pruning the backbone NN. 

For approach 2) class, more methods designed for RNNs are collected here, since it is not quite easy to do direct pruning upon RNN compared with pruning upon DNN and CNN. Tensor Train format is a way of building low-dim representation for RNN\,\cite{Tjandra2017}: for a $d$-dim tensor with $n$ possible values for each dimension index, they use the product of $d$ matrix (each slot has $n$ options of picking up a matrix) to represent each possible tensor elements. Assume the number of rows and columns for these matrices are $r$ (except for the first and the last matrix), then the number of elements needed to represent the original tensor size $n^d$ are now compressed to $\sim dnr^2$. There is another way to reducing the matrix representation by using block-circulant matrix\,\cite{wang2018}, where each $k\times k$ block in the original matrix are now represent by a $k\times 1$ vector circulant for $k$ times. Moreover, Fourier Transform of a circulant convolution will become a dot product, thus the computation can be speeded-up via FFT and IFFT. People also tried to simplify the macroscopic structure by inserting deep neural networks (DNN) into the microcosmic units. In \cite{Dai2018}, a DNN is inserted between the recurrent layer and the input (masking) layer for each gate in the LSTM to form a LSTM embedded with hidden layers (hLSTM). By modifying the microcosmic architecture (i.e. the LSTM cell), they are able to simplify the macroscopic architecture--say, reducing the number of LSTM layers--to achieve the more efficient setup (e.g. fewer number of total parameters, faster training speed, etc.). Furthermore, they can do traditional DNN pruning directly upon this newly added hidden layer. There are also some fundamentally different ways to do so such as binarized neural networks \cite{Courbariaux2016} where they can run the training and testing directly via the binary matrix multiplication in the GPU kernel.

\section{Contribution}
D.L. drew up the topic of this project. D.L. and N.S. obtained the data. M.Z. processed the data. D.L. designed the architecture of the original model. N.S. coded the original model. N.S. implemented the RNN pruning. D.L. implemented the DNN pruning and quantization. M.Z. coded the hLSTM class in Keras. N.S. and D.L. analyzed the behaviors of different kinds of model. M.Z. N.S. and D.L. all contributed to writing the final report.

\section{Acknowledgement}
We thank the authors in the artile ``Multitask Learning and Benchmarking with Clinical Time Series Data'' for providing the basic code for MIMIC-III data pre-processing on GitHub.

\end{document}